\documentclass[letterpaper]{article} 
\usepackage{aaai25}  
\usepackage{color,soul}
\usepackage{multirow}
\usepackage{booktabs}

\usepackage{times}  
\usepackage{helvet}  
\usepackage{courier}  
\usepackage[hyphens]{url}  
\usepackage{graphicx} 
\usepackage{natbib}  
\usepackage{caption} 
\usepackage{algorithm}
\usepackage{algorithmic}

\usepackage{xcolor}
\newcommand{\answerYes}[1]{\textcolor{blue}{#1}} 
\newcommand{\answerNo}[1]{\textcolor{teal}{#1}} 
\newcommand{\answerNA}[1]{\textcolor{gray}{#1}}

\usepackage{newfloat}
\usepackage{listings}
\DeclareCaptionStyle{ruled}{labelfont=normalfont,labelsep=colon,strut=off} 
\floatstyle{ruled}
\newfloat{listing}{tb}{lst}{}
\floatname{listing}{Listing}
\title{UKElectionNarratives: A Dataset of Misleading Narratives Surrounding Recent UK General Elections}
\author{
    Fatima Haouari\textsuperscript{\rm 1}, Carolina Scarton\textsuperscript{\rm 1}, Nicolò Faggiani\textsuperscript{\rm 2}, Nikolaos Nikolaidis\textsuperscript{\rm 3},\\ Bonka Kotseva\textsuperscript{\rm 4}, Ibrahim Abu Farha\textsuperscript{\rm 1}, Jens Linge\textsuperscript{\rm 5}, Kalina Bontcheva\textsuperscript{\rm 1}
}
\affiliations{
    \textsuperscript{\rm 1}Department of Computer Science, The University of Sheffield, Sheffield, UK\\
    \textsuperscript{\rm 2}Engineering S.p.A., Rome, Italy\\
     \textsuperscript{\rm 3}Athens University of Economics and Business, Athens, Greece\\
    \textsuperscript{\rm 4}Piksel S.r.l., Ispra, Italy\\
     \textsuperscript{\rm 5}European Commission, Joint Research Centre, Ispra, Italy\\
  {\{f.haouari, c.scarton, k.bontcheva\}@sheffield.ac.uk}

}

\usepackage{bibentry}

\begin{document}

\maketitle

\begin{abstract}
Misleading narratives play a crucial role in shaping public opinion during elections, as they can influence how voters perceive candidates and political parties. This entails the need to detect these narratives accurately. To address this, we introduce the first taxonomy of common misleading narratives that circulated during recent elections in Europe. Based on this taxonomy, we construct and analyse \textbf{UKElectionNarratives}: the first dataset of human-annotated misleading narratives which circulated during the UK General Elections in 2019 and 2024. We also benchmark Pre-trained and Large Language Models (focusing on GPT-4o),  studying their effectiveness in detecting election-related misleading narratives. 
Finally, we discuss potential use cases and make recommendations for future research directions using the proposed codebook and dataset.   
\end{abstract}
%

\section{Introduction}

Numerous misleading narratives tend to emerge during elections in Europe, often reflecting a mix of local and global concerns~\cite{ParliamentaryElections2019,EDMOElections2024}.
Specifically, this paper adopts the European Digital Media Observatory (EDMO) definition which describes misleading narratives as  \textit{``the clear message that emerges from a consistent set of contents that can be demonstrated as
false using the fact-checking methodology''}\cite{EDMOElections2024}. 

While each European country may encounter unique narratives shaped by its national political context, some shared misleading narratives tend to emerge as well. Common divisive ones tend to focus on the integrity of the electoral process, economic uncertainty, foreign interference, and social issues, such as gender dynamics, religious tensions, and immigration. These narratives exploit societal divisions and amplify existing tensions to influence public opinion, and frequently serve as focal points for disinformation campaigns~\cite{EDMOElections2024}. Therefore, a dataset of common misleading election narratives can provide a valuable, unique opportunity for researchers to develop innovative techniques for detecting, analysing, and countering narratives and disinformation campaigns on a large scale~\cite{ParliamentaryElections2019}.

This paper addresses this crucial research need to study misleading narratives created and spread on social media during elections. The main contributions of this work are:

\begin{table*}[!htb]
\scriptsize
    \centering

{\begin{tabular}{l c c c c c c c }\toprule
{\bf Dataset} & {\bf Source} & {\bf \# Instances}   &{\bf \# Narratives}& {\bf Topic}& {\bf Lang}\\\midrule
\bf CARDS~\cite{coan2021computer} &Articles&28,945&5/27&Climate Change&En\\
\bf Climate~\cite{piskorski2022exploring} &Articles&1,118&5/27&Climate Change&En\\

\bf COVID-19~\cite{kotseva2023trend} & Articles&58,625 &12/51&COVID-19&En\\
\bf COVID-19-Ger~\cite{heinrich-etal-2024-automatic} & Telegram&1,099&14&COVID-19&De\\

\bf DIPROMATS~\cite{fraile2024automatic} & Twitter&1,272/1,272&4/24&Narratives around COVID-19 by Authorities&En/Es\\
\bf TweetIntent@Crisis~\cite{ai2024tweetintent} &Twitter&3,691&2&Russia-Ukraine Crisis&En\\
\bf Climate\_ads~\cite{rowlands-etal-2024-predicting} &Facebook&1,330&7&Climate Change in US ads&En\\

\midrule
\textbf{UKElectionNarratives} &   Twitter     &2,000   &10/32     &UK Elections 2019/2024& En\\ 
\bottomrule
\end{tabular}}

    \caption{Comparison between \textbf{UKElectionNarratives} and existing \textit{public} datasets for detecting misleading narratives.}  
    \label{data_compare}
\end{table*}
\begin{enumerate}
\item The first public multi-level taxonomy of misleading narratives that emerged during recent elections in Europe.\footnote{\url{https://doi.org/10.5281/zenodo.15228071}} 
\item The first human-annotated social media dataset \textbf{UKElectionNarratives}\footnote{\url{https://doi.org/10.5281/zenodo.15228283}}  of misleading narratives that were spreading during the UK General Elections in 2019 and 2024.
\item Benchmarking results on the new \textbf{UKElectionNarratives} dataset, which explore the effectiveness of Pre-trained (PLMs) and Large Language Models (LLMs) (particularly, GPT-4o) in detecting misleading narratives. The source code of these experiments is also made available for reproducibility.\footnote{\url{https://github.com/GateNLP/UKElectionNarratives}}

\end{enumerate}

\section{Related Work}\label{RW}

This section discusses related work and gaps in the detection and discovery of misleading narratives, together with an examination of existing relevant datasets.

\subsection{Misleading Narratives Studies}
Research on misleading narratives spans various areas, employing diverse methodologies and datasets. In the health domain, \citet{ganti-etal-2023-narrative} re-annotate health misinformation datasets to examine whether they contain identifiable narratives.
Focusing on women's health, \citet{john2024online} investigate misleading narratives about reproductive health across social media platforms and websites. Their study identifies 112 distinct narratives, encompassing issues such as contraception, abortion, and fertility.

\citet{langguth2023coco} study stances toward 12 prevalent COVID-19 narratives on Twitter, emphasizing the role of social media in shaping public opinion. \citet{WICO} delve into conspiracy theories, focusing on COVID-19 related 5G claims and evaluating whether tweets expressed relevant allegations. Furthermore, \citet{introne2020mapping} explore anti-vaccination narratives in online discussion forums, offering insights into their evolution and influence. A comprehensive codebook for COVID-19 narratives for a period of three years was proposed by~\citet{kotseva2023trend}.

Several studies focus on narratives about climate change.~\citet{zhou2024large} explore the use of LLMs to extract and analyse the underlying narrative structures in English and Chinese news articles related to climate change. Additionally, \citet{coan2021computer} introduce a multi-level taxonomy designed to categorise and understand climate change narratives.

In the political domain, \citet{amanatullah2023tell} identify pro-Kremlin narratives related to the Russian-Ukraine war. \citet{moral2023assembling} and \citet{moral2024tale} adopt an unsupervised approach to identify strategic narratives during the COVID-19 pandemic, disseminated by Chinese and Russian official government and diplomatic accounts respectively. Conversely, \citet{fraile2024automatic} argue that unsupervised approaches may lead to biases and, instead, propose an annotation methodology for creating narrative classification datasets. 
Recently, \citet{sosnowski2024eu} propose a dataset for classifying narratives as either credible or disinformation adopting a human-in-the-loop approach. They study the effectiveness of different LLMs in detecting disinformation narratives. In contrast, this paper targets narratives surrounding recent elections in Europe, with UK Elections as a case study. 

\subsection{Existing Narratives Datasets}
As presented in Table.~\ref{data_compare}, most of the existing datasets focus on COVID-19~\cite{kotseva2023trend,heinrich-etal-2024-automatic} or climate change narratives~\cite{coan2021computer,piskorski2022exploring,rowlands-etal-2024-predicting}. For instance, COVID-19 narratives posted by authorities in Europe, US, China and Russia are studied by~\citet{fraile2024automatic}. In contrast,~\citet{ai2024tweetintent} create a dataset of narratives about the Russian-Ukraine war. 

However, to the best of our knowledge, there is no dataset to date which contains human-annotated narratives that spread during elections in Europe or, specifically, during UK General Elections. To bridge this gap, we introduce \textit{the first unified codebook} of misleading narratives that spread during Elections in Europe, and then validate this narrative taxonomy by creating \textit{the first dataset} with misleading narratives spread during two UK General Elections.  

\begin{figure*}[t]
\centering
\includegraphics[width=0.97\textwidth]{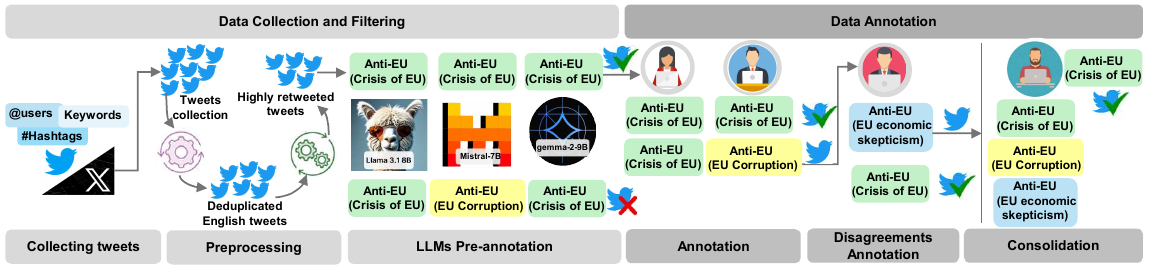} 
\caption{\textbf{UKElectionNarratives} construction approach. We first collect relevant tweets using a set of keywords, hashtags, and users. To select tweets with potential narratives, we select the highly retweeted tweets and those that 3 LLMs assigned identical narrative classes. The filtered tweets were then annotated by humans following a three-stage process.  }
\label{fig:data}
\end{figure*}

\section{Data Construction}\label{Data_construction}
Hereafter, we describe the details about our codebook and dataset.
As shown in Figure~\ref{fig:data}, the dataset was constructed by annotating a set of tweets following three main steps (1) collecting relevant tweets;  (2) data filtering; and (3) human annotation. 

\subsection{Election-based Misleading Narratives Codebook}\label{data:codebook} 

We developed a comprehensive codebook that systematically documents and defines the key misleading narratives that spread during elections in Europe, both at national and European Union level. Our literature-based methodology is based on thorough reviews of numerous relevant reports and scholarly publications on disinformation and election narratives surrounding the European Parliamentary Elections~\cite{ParliamentaryElections2019,ParliamentaryElections2019V4}. These reports  encompass a broad spectrum of European countries, including the United Kingdom, Spain, Germany, France, Italy, Poland, Hungary, Czechia, and Slovakia. 

In addition to the European elections level, we also analysed studies that focused on specific national elections, including the Swedish~\cite{colliver2018smearing}, Italian~\cite{alaphilippe2018developing}, German~\cite{applebaum2017make}, and French~\cite{ferrara2017disinformation} ones.  After analysing all these primary narratives, we engaged in  discussions and brainstorming sessions with a diverse team of social scientists and NLP experts, aiming to identify and refine a set of unified core narratives. This iterative process involved multiple rounds of evaluation and refinement to ensure the creation of a comprehensive and robust codebook.

Drawing on insights from political science literature, we refined and enhanced several narrative definitions in our codebook, ensuring greater alignment with theoretical frameworks to define specifically the \textit{Pro far-right}~\cite{rooduijn2024populist,mudde20242024,sibley2024behind,pirro2023far}, \textit{Pro far-left}~\cite{williams2018responding,rooduijn2024populist}, \textit{Anti-liberal}~\cite{coman2023anti}, and \textit{Anti-woke}~\cite{smith2023land} narratives which are classed under the \textit{Political hate and polarisation} super-narrative. 

Similar to other narrative studies~\cite{coan2021computer,kotseva2023trend,fraile2024automatic}, our codebook consists of general super-narratives and more specific narratives. For example, under the \textit{Anti-EU} super-narrative, which covers narratives that express opposition, scepticism, or criticism towards the European Union (EU), there are four  narratives namely \textit{EU economic scepticism}, \textit{Crisis of EU}, \textit{EU political interference}, and \textit{EU Corruption}. 

The final  codebook contains \textbf{10 super-narratives} and \textbf{32 narratives}. The full list of narratives and their descriptions is presented in the Appendix.

\subsection{Tweets Collection}
The dataset collection is focused specifically on narratives propagated on Twitter during the 2019 and 2024 UK General Elections, as a data-driven case study that validates the codebook. Thus, we collected tweets posted during both elections period as follows:
\begin{enumerate}
\item \textbf{UK General Election 2019 (UK\_GE\_2019):} Twitter data was collected in 2019 in real-time using the Twitter streaming API (V1.1). The data was collected using 33  hashtags (e.g., \#generalelection2019, \#borisjohnson) and 26 keywords (e.g., jeremy corbyn, caroline lucas). In addition we also included tweets from 18 political party accounts (e.g., @conservatives, @uklabour) where tweets posted by them or mentioning them were collected. Only tweets posted in November and December 2019 were included -- the election was held on 12 December 2019~\cite{prosser2021end}. This  collected dataset contains more than 60M tweets.
\item \textbf{UK General Election 2024 (UK\_GE\_2024):} Historical tweets were collected from the period of May 1\textsuperscript{st}, 2024 to July 4\textsuperscript{th}, 2024 -- the election was held on 4\textsuperscript{th} July 2024~\cite{duggan2024local}. Specifically tweets from 10 Labour user accounts (e.g., @SadiqKhan, @HackneyAbbott) and 6 Conservative user accounts (e.g., @RishiSunak, @SuellaBraverman) were collected.\footnote{We share all the hashtags, keywords, and user accounts used to collect our data.} 
Additionally, tweets mentioning these politicians by names are also included (e.g., Diane Abbott). This data consists of more than 500,000 tweets in total. 
\end{enumerate}

\subsection{Data Filtering}
Given that data annotation is costly, time-consuming and ambiguous, mainly for our case with a large amount of narratives, we considered a filtering pipeline. This filtering was essential for eliminating tweets without potential misleading narratives, thereby making the process more efficient and focused for subsequent human annotation. The filtering pipeline comprises the following steps:

\begin{enumerate}
\item \textbf{Selecting highly retweeted tweets:}
We selected the 20K top retweeted tweets from UK\_GE\_2019 and UK\_GE\_2024 (40K in total), and we excluded duplicates by keeping the tweet with the last timestamp.

\item \textbf{LLMs as pre-annotators:} Motivated by previous work that recently used LLMs for data filtering and annotation \cite{ai2024tweetintent,sosnowski2024eu}, we adopted three LLMs namely gemma 2~\cite{team2024gemma},\footnote{\url{https://huggingface.co/google/gemma-2-9b-it}}   Llama-3.1~\cite{touvron2023llama},\footnote{\url{https://huggingface.co/meta-llama/Llama-3.1-8B-Instruct}} and Mistral-7B~\cite{jiang2023mistral}\footnote{\url{https://huggingface.co/mistralai/Mistral-7B-Instruct-v0.2}} for pre-annotating the tweets into narratives, aiming to exclude tweets without potential narratives (see prompt in the Appendix). Specifically, for each tweet we prompted the LLMs to decide to which narrative class it belongs to. We then selected the tweets for which all three models agreed on their narrative labels, also excluding the tweets that were classified as not belonging to any narrative (i.e., labeled as \textit{None} by the three LLMs). We ended up with 3,281 and 2,695 tweets from UK\_GE\_2019 and UK\_GE\_2024 collections, respectively. To evaluate the effectiveness of this approach, we prompted the LLMs to label a set of 150 tweets, all of which had achieved full agreement between two human annotators and were validated by us during our initial pilot studies. Among these, the three LLMs assigned identical labels to 49 tweets. The Cohen's kappa~\cite{cohen1960coefficient} between the LLMs label and the humans label is 0.70, which indicates substantial agreement.

\item \textbf{Selecting an annotation sample:} From the data filtered in the previous step, we sampled 1,000 tweets from each election by selecting the top retweeted tweets from each narrative class. In total, we selected 2,000 tweets to be manually annotated.
\end{enumerate}

\subsection{Human Annotation}
To annotate our data, we hired 15 students whose majors are political sciences or journalism, and we conducted the annotation process using the collaborative web-based annotation tool Teamware~\cite{wilby2023gate}.\footnote{\url{https://gatenlp.github.io/gate-teamware/development/}}

Due to the complexity of the task and the fine-grained nature of our narratives classes, we focused heavily on delivering clear instructions while training the annotators. To achieve this, we conducted multiple pilot studies prior to the official annotation task, that helped in identifying the main challenges and difficulties faced by annotators.
As a result, our final annotation guidelines are the outcome of multiple rounds of refinement, each accompanied by internal discussions and analysis. 
Our pilot studies revealed that while the majority of tweets typically belong to a single narrative, a subset of tweets may encompass multiple narratives. To address this complexity, we introduced a confidence scoring system for annotators during the labeling process. Annotators were instructed to provide their confidence score for their assigned label on a scale of 1 to 5 (confidence scores descriptions are in the Appendix). If their confidence score fell below 4, they were required to assign a secondary label to the tweet. The inclusion of secondary labels and associated confidence scores have shown potential in improving the performance of some classification tasks as shown by~\citet{wu-etal-2023-dont}.

Before commencing the annotation process, all annotators attended an in-person training session. This training included a detailed explanation of the task, presentation of example tweets, and an opportunity for questions and discussion. Such preparatory training is needed to ensure annotators are capable to perform their task confidently and consistently. 
On average, annotators took an hour to label 50 tweets, earning an hourly wage of 17.2 GBP. To ensure the quality of our data, we designed an annotation process composed of \textit{three stages}, 
similar to previous work
~\cite{da2019fine,salman2023detecting}. The annotation process is as follows:

\paragraph{Stage 1 (annotation).} 
We adopted a batch approach \cite{maarouf2023hqp}, i.e., a group of annotators annotate a batch of data to prevent exhaustion. Thus, we split our data (2,000 tweets) into 20 batches with 100 tweets each, and each batch was annotated by two annotators independently. To ensure the quality of our annotations, annotators have to pass a qualification test in order to be considered qualified for the task. The qualification test was done by injecting an extra 25 gold tweets  (annotated during one of our pilot studies with full agreement of two annotators and checked by us) into the first batch annotated by each annotator. We then excluded the annotators who achieved Cohen’s kappa less than 0.4 on the gold set, and consequently, we re-annotated the batches annotated by those annotators. The overall Cohen’s kappa in \textit{Stage 1} is 0.34 which indicates fair agreement. The agreement score highlights the importance of the multi-stage annotation process, aligning with similar findings observed in existing fine-grained labelled datasets~\cite{da2019fine,salman2023detecting}.

\paragraph{Stage 2 (annotating tweets with disagreements).} A third annotator independently annotated the tweets with disagreements from \textit{Stage 1}. We then considered the majority voting to select the final narrative class. The tweets annotated by a third annotator have an average confidence score of 4, indicating a high level of certainty. This score shows that the third annotator was generally confident, with only minor uncertainty in a few cases.
\paragraph{Stage 3 (consolidation).} If after \textit{Stage 2} there is still disagreement between all annotators, we asked a fourth annotator (referred to as consolidator), who is a student in political sciences, to decide a final narrative class by checking the three previously assigned labels. The average confidence score for the consolidator is 4.34, which reflects the high level of confidence on the labels for this set.


\section{Data Overview}
In this section, we provide statistics about \textbf{UKElectionNarratives} and topic analysis.

\subsection{Data Statistics}\label{stats}

The most prevalent super-narratives in \textbf{UKElectionNarratives} are \textit{Distrust in institutions} and \textit{Distrust in democratic system} (see Table~\ref{tab:statistics}). However, the most dominant class is \textit{None} which we hypothesise to closely mirror the real-world scenario, where users comment on or question topics relevant to the elections without attempting to spread narratives. A similar pattern has been observed in datasets with stance towards rumors~\cite{gorrell-etal-2019-semeval,haouari2024authorities,zheng2022stanceosaurus}. It is worth noting, that although some tweets are labeled as \textit{None}, they may express other narratives not covered  in our codebook. Additionally,  similar to existing datasets with large number of classes, such as propaganda detection datasets~\cite{da2019fine,salman2023detecting}, the distribution of classes is unbalanced.
Finally, whilst \textbf{UKElectionNarratives} encompasses all 32 narratives  outlined in our codebook, the 2019 and 2024 elections data covers only 30 narratives and 27 narratives, respectivelly. We present example tweets from our dataset in Table~\ref{fig:example} (in Appendix).
\begin{table}[!htb]
\centering
\small
    \begin{tabular}{lrl}
    \toprule
      \textbf{Distrust in institutions} & 305 &(15.25\%) \\
              Failed state & 64 &\\
        Criticism of national policies  & 241  &  \\

     \midrule
      \textbf{Distrust in democratic system} & 191 &(9.55\%) \\
         Elections are rigged & 71 & \\
        Anti-Political system & 18 & \\
        Immigrants right to vote &1 & \\
        Anti-Media & 101& \\

    \midrule
       \textbf{Political hate and polarisation} & 135 &(6.75\%) \\
              Pro far-left &56 &\\
        Pro far-right  & 54  &  \\
        Anti-liberal &17&\\
        Anti-woke & 8 & \\

          \midrule
        \textbf{Anti-EU} &82&(4.1\%) \\
        EU economic scepticism & 19 &\\
        Crisis of EU   & 25 &  \\
       EU political interference & 33 & \\
       EU corruption & 5 &\\

    \midrule
       \textbf{Gender-related} & 177&(8.85\%) \\
         Language-related & 7 &\\
        LGBTQ+-related & 169 &\\
        Demographic narratives &1 &\\
            \midrule
      \textbf{Religion-related} & 109 &(5.45\%) \\
         Anti-Islam & 101 & \\
        Anti-Semitic conspiracy theories & 5 & \\
        Interference with states' affairs & 3& \\
        \midrule
        \textbf{Migration-related} & 91 &(4.55\%) \\
         Migrants societal threat & 91& \\
    \midrule
    \textbf{Ethnicity-related} & 23 &(1.15\%) \\
         Association to political affiliation & 7 & \\
        Ethnic generalisation &4 & \\
        Ethnic offensive language & 6 &\\
       Threat to population narratives & 6 & \\
    \midrule
       
        \textbf{Geopolitics} &90 &(4.5\%) \\
         Pro-Russia & 5 & \\
        Foreign interference & 81 & \\
        Anti-international institutions &4 & \\
    \midrule
     \textbf{Anti-Elites} &16 &(0.8\%) \\
         Soros & 6 & \\
        World Economic Forum / Great Reset & 4 & \\
        Antisemitism & 2 & \\
       Green Agenda &4 & \\

    \midrule

     \textbf{None} & 781 &(39.05\%) \\
    \midrule
    \textbf{Total} &2000&\\
    \bottomrule
\end{tabular}%
\caption{\textbf{UKElectionNarratives} statistics. \textbf{Super-narratives} are in bold. }
\label{tab:statistics}
\end{table}

\begin{table*}[!htb]
  \scriptsize
  \centering
  \label{tab:topics}
  \begin{tabular}{p{0.000001cm}cp{7.85cm}p{7.58cm}}
    \toprule
    &\bf Topic&\bf GPT4o Topic Label (\#Tweets)& \bf Most Prevalent Super-narrative (Narrative) [\%Tweets]\\
    \midrule
    \multirow{10}{*}{\rotatebox[origin=c]{90}{\textbf{UK Election 2019}}} & (1)& Jeremy Corbyn and the Labour Party (141)& None [69.5\%] / Political hate and polarisation (Pro far-left) [12.1\%] \\
    &(2) & UK and EU Relations (136)& Distrust in institutions (Criticism of national policies) [23.5\%] \\
    &(3)  & Islamophobia and Jihadist Attacks in Politics (119)&None [45.4\%] / Religion-related (Anti-Islam) [29.4\%]\\ 
     &(4) &Voter ID and Election Integrity Concerns (117)&Distrust in democratic system (Elections are rigged) [41\%]  \\
     &(5)  &Suppression of Russian Interference Report by Boris Johnson's Government (80) &Geopolitics (Foreign interference) [75\%]\\
     &(6)& Antisemitism and Politics (61)& None [77\%] / Distrust in democratic system (Anti-Media) [\%9.8]\\
     &(7) &  Trans Rights and Issues (59)&Gender-related (LGBTQ+-related) [83.1\%]\\
   &(8) &NHS and Healthcare Privatization Concerns (58)&Distrust in institutions (Criticism of national policies) [51.7\%]\\
    &(9) &Immigration Policies and Impact on Public Services (55)&Migration-related (Migrants societal threat) [49.1\%]\\
   \midrule
    \multirow{10}{*}{\rotatebox[origin=c]{90}{\textbf{UK Election 2024}}} & (1)& Criticism of Rishi Sunak's Leadership and Statements (130) & None [40\%] / Distrust in institutions (Criticism of national policies) [27.1\%]\\
    &(2)& UK Migrant Policy and Labour's Asylum Seeker Plan (112)&Migration-related (Migrants societal threat) [46.4\%] \\
    &(3)& UK Politics, Israel Lobby, and Antisemitism Debate (96)&None [65.6\%] / Distrust in institutions (Criticism of national policies) [20.8\%]\\ 
     &(4)& Trans Rights and Gender Ideology Debate in UK Politics (95) &Gender-related (LGBTQ+-related) [86.3\%]\\
     &(5) &Criticism of Keir Starmer and the Labour Party (95)&None [66.3\%] / Distrust in institutions (Criticism of national policies) [8.4\%]\\
     &(6) &Controversy Surrounding Sadiq Khan's Mayoralty of London (90)&Religion-related (Anti-Islam) [45.6\%] \\
     &(7) &UK Economy and EU Relations Under Keir Starmer (57) & None [43.9\%] / Distrust in institutions (Criticism of national policies) [24.6\%]\\
   &(8) &Diane Abbott and Her Treatment by the Labour Party (41)& None [90.2\%] / Distrust in institutions (Criticism of national policies) [4.88\%]\\
   &(9) &Labour Party Call for Change in Britain (36)& None [88.88\%] / Political hate and polarisation (Pro far-left) [5.56\%]\\
     \bottomrule
    \end{tabular}
  \caption{Detected topics in \textbf{UKElectionNarratives} using BERTopic and GPT4o, and the most prevalent super-narrative (narrative) for each topic based on our annotated data. We also present the second most prevalent if the first is the \textit{None} class.}    
   \label{tab:topics}
\end{table*}

\subsection{Topic Modeling Analysis}
To analyse the key topics discussed in tweets during each of the UK elections, we conducted a topic modeling analysis using BERTopic~\cite{grootendorst2022bertopic}. BERTopic has recently gained popularity~\cite{hellwig2024exploring,zain2024revealing} due to its ability to capture semantic relationships between words through document embeddings, which leads to the generation of more insightful topics.
For tweet embeddings, we utilised the base GTE model\footnote{\url{https://huggingface.co/thenlper/gte-base}} \cite{li2023towards}. We then applied UMAP~\cite{mcinnes2018umap} for dimensionality reduction, and used HDBSCAN~\cite{mcinnes2017hdbscan} for clustering. By setting the minimum number of tweets per topic to 25, we successfully identified 9 distinct topics for each election. 

As highlighted by~\citet{alammar2024hands}, LLMs have the potential to achieve remarkable outcomes when leveraged for labeling the topics identified by BERTopic. 
In our study, we employed GPT-4o~\cite{hurst2024gpt} for generating topic labels based on the keywords and the most representative tweets extracted by BERTopic, following the official prompt guidelines recommended by the BERTopic author. In our prompt (presented in the Appendix), we set the number of representative tweets as 10.
\subsubsection{Detected Topics}
Table~\ref{tab:topics} presents the topics identified for each election. Additionally, we show the most dominant super-narrative (narrative) class for each topic, illustrating their alignment with the topic labels.

For the 2019 election, we observe that the labels for topics 4, 5, 7, 8, and 9 align precisely with their most prevalent super-narrative (narrative). For topic 2, while the most prevalent super-narrative (narrative) is \textit{Distrust in institutions (Criticism of national policies)}, at the super-narrative level alone, \textit{Anti-EU} emerges as the most prevalent, accounting for 46.3\% of the tweets associated with this topic. Differently, for topics 1, 3, and 6, the most dominant narrative class is \textit{None}. However, examining the second most dominant class for these topics reveals alignment with the topic labels. For topic 1, the second most prevalent class is \textit{Political hate and polarisation (Pro far-left)}. We believe this aligns with the topic label, as in the UK, discussions around left-wing populism have largely centered on Jeremy Corbyn's leadership of the Labour Party~\cite{rios2022between}. Similarly, for topic 3, the second most dominant class is \textit{Religion-related (Anti-Islam)} which aligns with the Islamophobia topic. Lastly, for topic 6, although the second most prevalent class is \textit{Distrust in democratic system (Anti-Media)}, a manual examination of the tweets for this topic showed that they all discuss antisemitism as shown in this tweet example ``\textit{Jews, antisemitism and Labour – a letter to the BBC about the systematic bias in their reporting of this issue and their continued pattern of accepting assertion as fact and not allowing contrary opinions a hearing opinion.}''. Moreover, we found that all the tweets labeled as \textit{Religion-related  (Anti-Semitic conspiracy theories)}  are within the tweets for this topic.

For the 2024 election, we observe that the labels for topics 2 and 4 align perfectly with their most dominant super-narrative (narrative). For topic 6, the most prevalent narrative is \textit{Religion-related (Anti-Islam)}. The reason is that Sadiq Khan is London's first Muslim mayor~\cite{zheng2021islamophobia} and during the 2024 election, he encountered significant Islamophobic rhetoric from certain politicians. Notably, Conservative MP Lee Anderson claimed that ``Islamists'' had ``taken control'' of Khan~\cite{lucas2024liz}. Looking at the other topics where the dominant class is \textit{None}, we observe that topics 3, 5, 7, and 8 are all relevant to \textit{Distrust in institutions (Criticism of national policies)} narrative. Additionally, we found that for topic 3, all the tweets labelled as \textit{Anti-Elites (Antisemitism)} and \textit{Religion-related  (Anti-Semitic conspiracy theories)} are within the tweets for this topic. Finally, for topic 9, the second most prevalent class is \textit{Political hate and polarisation (Pro far-left)}, which is highly relevant to a topic about the Labour Party, positioned on the left of the UK political spectrum~\cite{grant2024new}.

\subsubsection{Topic Similarity Between Both Elections}
To demonstrate the similarities between narratives across different elections, we identified the overlapping topics between the two elections by leveraging their respective topic embeddings. To quantify the alignment between these topics, we calculated their cosine similarity, providing a measure of how closely related the narratives are. For visualization, we used a cosine similarity matrix, which as presented in 
Figure~\ref{fig:topics_heatmap} (in the Appendix), maps the degree of similarity between narratives across the two elections. This approach highlights how some misleading narratives can persist across different elections such as those concerning EU relations, Immigration, Gender, Antisemitism, and Islamophobia. This also demonstrates how these issues continue to be manipulated and distorted over time, shaping political discourse.

\begin{table*}[!htb]
\small
\centering
  \begin{tabular}{c|ccc|ccc}
    \toprule
    &\multicolumn{3}{c|}{\textbf{Development Set}} &\multicolumn{3}{c}{\textbf{Test Set}} \\ 
      & Macro-$F_1$ &Macro-Rec&Macro-Prec& Macro-$F_1$ &Macro-Rec&Macro-Prec\\
    \midrule

    Majority Classifier&0.018&0.032&0.012&0.017&0.030&0.012\\ 
   Random Classifier& 0.019&0.018&0.032&0.021&0.073&0.039\\ \midrule
   RoBERTa-base&0.227&0.235&0.238&0.216&0.220&0.233\\
   RoBERTa-large&0.266&0.291&0.263&0.226&0.248&0.215\\
   \midrule
      GPT-4o\textsubscript{(zero-shot)}&0.423&0.521&0.426&0.444&0.534&0.460\\ 
    GPT-4o\textsubscript{(zero-shot+Narrative descriptions)}&\underline{0.435}&\underline{0.545}& \underline{0.447}&\underline{0.479}&\underline{0.554}& \bf 0.509 \\ 
   \midrule
    GPT-4o\textsubscript{(few-shot)}&0.429
&0.513&\bf 0.448&0.421&0.480&0.447\\ 
   GPT-4o\textsubscript{(few-shot+Narrative descriptions)}&\bf 0.445&\bf 0.560&0.438&\bf 0.488&\bf 0.555&\underline{0.504}  \\ 
    \bottomrule
  \end{tabular}
 \caption{Benchmarking results on the \textbf{UKElectionNarratives} development and test sets.} 
   \label{tab:results}
\end{table*}
\section{Experimental Setup}
This section presents the experimental setup to showcase the use of \textbf{UKElectionNarratives} in a narratives classification benchmark.

\subsection{Narrative Detection Models}

We adopted three categories of models for benchmarking:
\paragraph{Basic Models} We considered two dummy baselines:\footnote{\url{https://scikit-learn.org/stable/modules/generated/sklearn.dummy.DummyClassifier.html}} (1) \textit{Majority Classifier}: that always predicts the most frequent class in the training data, and (2) \textit{Random Classifier}: that randomly generates predictions with equal probability for each unique class observed in the training data.

\paragraph{Pre-trained Language Models (PLMs)} We fine-tuned the base\footnote{\url{https://huggingface.co/FacebookAI/roberta-base}} and large\footnote{\url{https://huggingface.co/FacebookAI/roberta-large}} RoBERTa~\cite{liu2019roberta} models to evaluate their performance on detecting misleading narratives. We fine-tuned both models for 5 epochs with a batch of size 16, experimenting with different learning rates [1e-5, 2e-5, 3e-5], and dropout values [0.1, 0.2, 0.3]. We then selected the best model based on Macro-$F_1$ on the dev set.
\paragraph{Large Language Models (LLMs)} We adopted OpenAI GPT4o~\cite{hurst2024gpt}, which has demonstrated high performance in detecting disinformation narratives compared to multiple LLMs~\cite{sosnowski2024eu}. Our experiments included both \textit{zero-shot} and \textit{few-shot} setups. We optimized the prompts on the development set to identify the best-performing prompts, which were then used to evaluate the test set. To ensure reliability, we executed each prompt three times and reported the average performance. We present our prompts under all setups in Appendix.

\subsection{Data Splits}
We applied stratified sampling~\cite{sechidis2011stratification} to split the data into training (70\%), development (dev) (10\%), and test (20\%) sets. For the narratives with 1 or 2 examples (3 narrative classes), no examples were included in the training set. The labels distribution on each set is presented in Table~\ref{tab:labels_distribution} in the Appendix. 

\section{Experimental Evaluation}\label{result}


 

\paragraph{GPT-4o versus RoBERTa} As presented in Table~\ref{tab:results}, GPT-4o significantly outperforms both RoBERTa models across all evaluation measures. This improvement can be attributed on the one hand to GPT-4o extensive training data covering relevant topics. On the other hand, our data is unbalanced with some narratives having a small number of examples, which is challenging for PLMs. Moreover, three narrative classes are missing from the training data which poses a significant limitation as PLMs are unable to learn patterns for the missing classes. 

\paragraph{GPT-4o zero-shot versus few-shot setup}
For the \textit{few-shot} setup, we provided a tweet example for every narrative in the training data. As presented, in Table~\ref{tab:results}, the results show that providing tweet examples, slightly improved the performance on the dev set in terms of Macro-$F_1$ and precision but degraded the recall. For the test set, the \textit{zero-shot} setup showed better performance. However, we believe more investigation exploring other prompt engineering techniques may further improve the results~\cite{sahoo2024systematic}.

\paragraph{Narrative descriptions in GPT-4o prompt}
We included the description for each narrative, as per our codebook, in the prompt to determine whether this additional information would help the model better understand the narrative classes and distinguish between them more effectively. The results show that, for both the \textit{zero-shot} and \textit{few-shot} setups, including the narrative descriptions enhanced the model's performance compared to simply listing the narrative classes.  Finally, combining the narrative descriptions and tweets examples resulted in the best overall performance in both dev and test sets. We argue that, when considering the trade-off between cost and effectiveness, the \textit{zero-shot+Narrative descriptions} setup can be sufficient, as the improvement in performance is relatively modest compared to the cost of including tweet examples (30 tweet examples in our setup).


\section{Potential Use Cases}
In this section, we outline a few potential use cases for our \textit{codebook} and \textit{dataset}, illustrating how they can be leveraged for various tasks and challenges.
\paragraph{Constructing novel datasets of misleading narratives spread during elections in Europe} Our multi-level codebook can aid in annotating new datasets for detecting misleading narratives that are propagated during elections in European countries other than the UK. Furthermore, the codebook can be easily adapted by extending or replacing certain narratives to better suit the context of a specific country or to address emerging misleading narratives in future elections. This flexibility ensures that the codebook remains relevant and effective in representing new narratives as they arise.
\paragraph{Enabling automatic annotation}
Despite the limited number of tweets in \textbf{UKElectionNarratives}, it has the potential to significantly enhance the process of automatic labelling for a much larger volume of tweets, as done by~\citet{ai2024tweetintent}. By leveraging our dataset, models that are capable of efficiently labelling additional tweets with high accuracy can be developed. This approach not only simplifies the annotation process but also expands the scope of analysis. 
Domain adaptation techniques can also be explored to generalise models developed with our dataset to elections in other countries and even to other domains (e.g. health-related misinformation).

\paragraph{Misinformation analysis} Given the overwhelming volume of information that circulated about the UK General Elections, numerous misleading narratives rapidly spread and gained significant attention from the public~\cite{vaccari2023campaign,gaber2022strategic}. The accelerated distribution of misinformation poses a serious challenge, as it can quickly shape public perceptions. 
In turn, this may result in negative outcomes, including the loss of trust in political institutions, the spread of confusion among voters, and potentially harmful effects on the democratic process itself. Our dataset can play a crucial role in supporting the study of misinformation detection and verification during this period, providing insights for countering misleading narratives in future events.

\paragraph{Analysing election discourse for research and media literacy} Elections spark widespread discussions on social media, where individuals express opinions on policies, candidates, and political events. These conversations, often driven by strong emotions, provide valuable data for analysing public sentiment~\cite{rita2023social}, political stance~\cite{muller2022real}, hate speech~\cite{hate}, and other forms of offensive or harmful language~\cite{weissenbacher-kruschwitz-2024-analyzing}. 
The social media discussions in \textbf{UKElectionNarratives} can support research on political communication and digital discourse, enabling scholars to study narrative patterns, assess online engagement, and develop tools that promote media literacy, fostering informed public discourse.
\section{Limitations}
This study presents a codebook for the main narratives surrounding the European Elections; however, it may not cover certain country-specific narratives. Therefore, the codebook should be expanded to accommodate the context of specific European countries or incorporate new emerging misleading narratives in the future.

One limitation of our dataset is the potential for annotator bias. Despite efforts to ensure consistency through guidelines and training, individual annotators may interpret narratives subjectively, leading to variations in classification. Additionally, due to time and cost constraints, the third stage of annotation involved a single consolidator for each tweet. A better approach would be having two consolidators to discuss and decide a final narrative class. We also acknowledge that the UK Election domain is limited and will not reflect all narratives that appear in other European countries. Furthermore, the size of the dataset presented herein is relatively small and certain narratives have very few examples.

In terms of experiments, further exploration is required to evaluate additional LLMs and diverse prompt engineering techniques. Moreover, techniques to mitigate the class imbalance should be investigated, alongside data augmentation methods to improve the performance of the models.





\section{Conclusion and Future Work}
In this paper, we introduced a multi-level codebook with a taxonomy of the main misleading narratives spread during elections in Europe. Adopting this taxonomy, we created \textbf{UKElectionNarratives}, the first dataset annotated with misleading narratives spread during the 2019 and 2024 UK General Elections. We performed a topic modelling analysis on this data, and demonstrated how topics align with annotated narratives. Moreover, we presented benchmarking results on our dataset, assessing PLMs and LLMs performance. We found that GPT-4o significantly outperforms RoBERTa-based models. However, further exploration is needed to assess additional LLMs and experiment with various prompt engineering techniques. Additionally, approaches to mitigate class imbalance should be examined, in conjunction with data augmentation methods, to enhance the models performance. As a future work, we plan to augment our dataset adopting different techniques including a human-in-the-loop approach, automatic data annotation, and generating synthetic data using our proposed codebook and novel dataset. Moreover, we aim to propose more robust models tailored for detecting misleading narratives.

\section{Acknowledgements}
This work is supported by the UK’s innovation agency (InnovateUK) grant number 10039039 (approved under the Horizon Europe Programme as VIGILANT, EU grant agreement number 101073921) (\url{https://www.vigilantproject.eu}).

\bibliography{main}

\section{Paper Checklist}

\begin{enumerate}

\item For most authors...
\begin{enumerate}
    \item  Would answering this research question advance science without violating social contracts, such as violating privacy norms, perpetuating unfair profiling, exacerbating the socio-economic divide, or implying disrespect to societies or cultures?
    \answerYes{Yes. Our dataset and codebook provide unique resources for the study of misinformation narratives in (European) elections. By creating a codebook grounded on the literature, we believe to have minimised biases inherently present in data-driven approaches. The datasest complies with the ethics standards of the University of Sheffield Research and Ethics (UREC) guidelines\footnote{\url{https://www.sheffield.ac.uk/rpi/ethics-integrity}} (in particular, Ethics note number 14),\footnote{\url{https://www.sheffield.ac.uk/media/29459/download?attachment}} making available only anonymised data and focusing on general misinformation narratives, instead of specific profiles. By collecting data focusing only on generally used hashtags and public figures, we also aimed to avoid including biased and/or personal data about individuals.}
  \item Do your main claims in the abstract and introduction accurately reflect the paper's contributions and scope?
    \answerYes{Yes, the abstract and introduction contain an accurate description of the contributions and scope of our paper.}
   \item Do you clarify how the proposed methodological approach is appropriate for the claims made? 
    \answerYes{Yes. The codebook is created by focusing on a extensive literature review, basing the narrative definitions in such previous work. The dataset is created using a well-established framework for social media data collection, focusing on keywords (hashtags) and public figures monitoring. Data annotation is conducted by using a robust three-stage methodology, aiming to remove biases and solve disagreements. The methodology for benchmarking the dataset also follows the state-of-the-art for the field.}
   \item Do you clarify what are possible artifacts in the data used, given population-specific distributions?
    \answerNo{No, because the dataset does not contain population-specific features (apart from generally being only UK centred). See our Ethics Statement for more details.}
  \item Did you describe the limitations of your work?
    \answerYes{Yes, a limitations section is explicitly added in the paper.}
  \item Did you discuss any potential negative societal impacts of your work?
    \answerYes{Yes. Our Ethics Statement presents a discussion about our analysis being only on aggregated data, completely avoiding drawing conclusions about individuals. }
      \item Did you discuss any potential misuse of your work?
    \answerYes{Yes. The Limitations section discuss the current limitations of our work highlighting cases of potential misuse.}
    \item Did you describe steps taken to prevent or mitigate potential negative outcomes of the research, such as data and model documentation, data anonymization, responsible release, access control, and the reproducibility of findings?
    \answerYes{Yes. Our data sharing process is restricted to only sharing tweet IDs and the dataset license only allows research work and requests that modifications to this dataset to be released under the same license. Our code is also released to ensure reproducibility.}
  \item Have you read the ethics review guidelines and ensured that your paper conforms to them?
    \answerYes{Yes. We have read the guidelines and ensured our paper conforms to them.}
\end{enumerate}

\item Additionally, if your study involves hypotheses testing...
\begin{enumerate}
  \item Did you clearly state the assumptions underlying all theoretical results?
    \answerNA{NA}
  \item Have you provided justifications for all theoretical results?
    \answerNA{NA}
  \item Did you discuss competing hypotheses or theories that might challenge or complement your theoretical results?
    \answerNA{NA}
  \item Have you considered alternative mechanisms or explanations that might account for the same outcomes observed in your study?
    \answerNA{NA}
  \item Did you address potential biases or limitations in your theoretical framework?
    \answerNA{NA}
  \item Have you related your theoretical results to the existing literature in social science?
    \answerNA{NA}
  \item Did you discuss the implications of your theoretical results for policy, practice, or further research in the social science domain?
    \answerNA{NA}
\end{enumerate}

\item Additionally, if you are including theoretical proofs...
\begin{enumerate}
  \item Did you state the full set of assumptions of all theoretical results?
    \answerNA{NA}
	\item Did you include complete proofs of all theoretical results?
    \answerNA{NA}
\end{enumerate}

\item Additionally, if you ran machine learning experiments...
\begin{enumerate}
  \item Did you include the code, data, and instructions needed to reproduce the main experimental results (either in the supplemental material or as a URL)?
    \answerYes{Yes. We share both code and prompts used in our experiments. Data is shared as described in the Ethics Statement (under a share-alike CC license for research purposes only).}
  \item Did you specify all the training details (e.g., data splits, hyperparameters, how they were chosen)?
    \answerYes{Yes. Model details are presented in the paper and available in the GitHub repository.}
     \item Did you report error bars (e.g., with respect to the random seed after running experiments multiple times)?
    \answerNA{NA}
	\item Did you include the total amount of compute and the type of resources used (e.g., type of GPUs, internal cluster, or cloud provider)?
    \answerYes{Yes. We describe the resources used in the Appendix. However, it is worth noting that for GPT-4o we used the OpenAI API, which does not explicitly describe the hardwared used.}
     \item Do you justify how the proposed evaluation is sufficient and appropriate to the claims made? 
    \answerYes{Yes. Our benchmarking follows established evaluation metrics from previous work. }
     \item Do you discuss what is ``the cost`` of misclassification and fault (in)tolerance?
    \answerNo{No, because we do not perform an error analysis which is beyond the scope of this paper. However, we provide details in the Limitations section that can be used to mitigate misclassification in future work.}
  
\end{enumerate}

\item Additionally, if you are using existing assets (e.g., code, data, models) or curating/releasing new assets, \textbf{without compromising anonymity}...
\begin{enumerate}
  \item If your work uses existing assets, did you cite the creators?
    \answerNA{NA}
  \item Did you mention the license of the assets?
    \answerYes{Yes. The license of our dataset is CC-BY-NC-SA. }
  \item Did you include any new assets in the supplemental material or as a URL?
    \answerYes{Yes. A sample of our dataset is added as supplementary material and the codebook is added as part of the Appendix.}
  \item Did you discuss whether and how consent was obtained from people whose data you're using/curating?
    \answerYes{Yes. Our annotators consented to the use of their annotations in our research. Although it is not possible to get consent from social media users, we follow the GDPR and UREC guidelines for the use of social media posts.}
  \item Did you discuss whether the data you are using/curating contains personally identifiable information or offensive content?
    \answerYes{Yes. In our ethics statement we mention potential harms to annotators regarding the nature of social media data. However, as previous work have identified, toxic content and hate speech are not as frequent as other content, which reduces the chances of exposure to such data.}
\item If you are curating or releasing new datasets, did you discuss how you intend to make your datasets FAIR (see \citet{fair})?
\answerYes{Yes. We made our dataset available through Zenodo and provided the details of its license and permitted usage in the Ethics Statement.}
\item If you are curating or releasing new datasets, did you create a Datasheet for the Dataset (see \citet{gebru2021datasheets})? 
\answerNo{No. However, we provide all details regarding data collection, annotation and release in the paper (therefore, the information that should appear in the datasheet are already detailed in the paper). We also clearly define the motivation behind data creation and overall research topic.}
\end{enumerate}

\item Additionally, if you used crowdsourcing or conducted research with human subjects, \textbf{without compromising anonymity}...
\begin{enumerate}
  \item Did you include the full text of instructions given to participants and screenshots?
    \answerYes{Yes. We added the instructions given to annotators in the Appendix.}
  \item Did you describe any potential participant risks, with mentions of Institutional Review Board (IRB) approvals?
    \answerYes{Yes. We follow the UREC guidelines and clearly inform participants of potential risks in participating in an annotation task involving social media (e.g. the exposure to toxic content). Participants were made aware of these risks and were given time to read the details of consent form.}
  \item Did you include the estimated hourly wage paid to participants and the total amount spent on participant compensation?
    \answerYes{Yes. This is described in the paper under the Human Annotation subsection.}
   \item Did you discuss how data is stored, shared, and deidentified?
   \answerYes{Yes. All data is securely stored in our encrypted servers. The data shared with annotators is done via our Teamware tool, that also stores all data in our encrypted servers. Annotators data (e.g. e-mail) are stored separately from the annotations and researchers only have access to annotators' IDs and annotators (i.e. no access to personal information is available). The dataset is made available for research without any identification to the annotators. The data from social media is also processed and annotated following the UREC guidelines as previously stated.}
\end{enumerate}

\end{enumerate}

\section{Ethics Statement}
\paragraph{Data Collection} Our data collection complies with social platform T\&Cs and the UK Data Protection Act.\footnote{https://www.gov.uk/data-protection}

\paragraph{Data Annotation}
The annotation task has been approved by the University Research and Ethics Committee (UREC) with reference number 059166.\footnote{\url{https://www.sheffield.ac.uk/rpi/ethics-integrity}} Annotators were recruited through an internal process facilitated by the University. They were informed of potential harms related to their participation and given training for data annotation. They were informed that they could leave the task at any time, without any negative consequences to them and they also signed a consent form. They were informed that the data could contain toxic language and hate speech and they were allowed to skip tweets if they do not feel comfortable. Our approach involving three stages also mitigate annotators' biases regarding subjective aspects (e.g. political and cultural views). 

\paragraph{Data Release} Adhering to X T\&Cs,\footnote{\url{https://developer.x.com/en/more/developer-terms/agreement-and-policy}} we only release tweet IDs together with their respective annotations under a CC-BY-NC-SA 4.0 license.\footnote{\url{https://creativecommons.org/licenses/by-nc-sa/4.0/deed.en}} The dataset is available in the following link: https://doi.org/10.5281/zenodo.15228283

\paragraph{Data Analysis} Although we use anonymised data in our experiments, we are aware of the potential risks of using social media data and profiling. Therefore, our analysis are always about aggregated results and we do not have or provide access to demographic information.

\appendix
\section*{Appendix}

\section{Narrative Descriptions in Our Codebook}
\label{apx:codebook}

\begin{enumerate}
    \item \textbf{Anti-EU (EU economic scepticism):} Narratives criticising EU's economic policies and how they affect local economies.
    \item \textbf{Anti-EU (Crisis of EU):} Narratives expressing scepticism or concern about the EU, focusing on alleged broader issues affecting the union's legitimacy and effectiveness.
    \item \textbf{Anti-EU (EU political interference):}  Narratives alleging that the EU is interfering or manipulating local politics and specific policy areas, including food, health, environmental, and migration policies, sparking controversy and debate over EU influence on national sovereignty.
    \item \textbf{Anti-EU (EU Corruption):} Narratives claiming corruption within the European Union and its institutions.
    \item \textbf{Political hate and polarisation (Pro far-left):} Refers to the manipulative narratives that fundamentally reject the existing socio-economic structure of contemporary capitalism. These narratives advocate for an alternative economic and power structure based on principles of economic justice and social equality. They view economic inequality as a core issue within the current political and social arrangements and propose a major redistribution of resources from existing political elites. Additionally, they support a state role in controlling the economy. These narratives target both ``radical left'' who accepts democracy but favors ``workers’ democracy'' and the direct participation of labor in the management of the economy, and the ``extreme left'' emphasizes both parliamentary, as well as the extra-parliamentary struggle against globalized capitalism, and sees no place for free-market enterprise~\cite{williams2018responding,rooduijn2024populist}.
    \item \textbf{Political hate and polarisation (Pro far-right):} Refers to the manipulative narratives that exploit concerns about immigration, national security, and the loss of control over domestic affairs to prioritize the interests of native citizens. These narratives advocate for strict societal authority and order, often framing these issues in a way that appeals to fear and insecurity, thereby promoting divisive and exclusionary policies~\cite{rooduijn2024populist,mudde20242024,sibley2024behind,pirro2023far}.
  \item \textbf{Political hate and polarisation (Anti-liberal):} Refers to the manipulative narratives that reject key principles of liberalism—political, economic, and cultural—and seek to reshape politics and societal norms. These narratives blame political and economic crises on liberal policies, particularly viewing the process of EU integration as a source of domestic issues. They advocate for a political community that promotes a narrow, shared cultural and moral identity, often framing these issues in a way that appeals to fear and exclusionary instincts~\cite{coman2023anti}.
\item \textbf{Political hate and polarisation (Anti-woke):} Narratives that frame efforts to address social inequalities—such as racism, sexism, and anti-LGBTQ+ discrimination—as exaggerated, illegitimate, or a threat to societal norms. These narratives often shift focus away from marginalized groups and instead position privileged groups, such as wealthy individuals, as victims of ``wokeness'' or cultural shifts. They emphasize cultural divisions, dismiss systemic inequalities as non-issues, and portray initiatives promoting minority rights or inclusion as oppressive, unnecessary, or harmful~\cite{smith2023land}. 
\item \textbf{Religion-related (Anti-Islam):} Criticising Islam and portraying it as a threat to national security and cultural identity. Also, claiming that Muslim communities pose a risk to public safety and that the spread of Islam in Europe could lead to the destruction of traditional Western values and cultural heritage.
\item \textbf{Religion-related (Anti-Semitic conspiracy theories):} Narratives promoting conspiracy theories that allege Jewish people or organisations are responsible for major global events or issues.
\item \textbf{Religion-related (Interference with states' affairs):} Narratives of foreign religious authorities interfering in a country's internal political processes.

\item \textbf{Gender-related (Language-related):} Narratives that claim certain gender terminologies or vocabularies are being imposed on individuals or communities. Controversy over language use and inclusivity.

\item \textbf{Gender-related (LGBTQ+-related):} Narratives related to the LGBTQ+ community. Those include discussions around issues of acceptance, rights, discrimination, etc.
\item \textbf{Gender-related (Demographic narratives):} Narratives that argue against masturbation or the 'gender propaganda' and claiming that they promote infertility and population decline.
\item \textbf{Ethnicity-related (Association to political affiliation):} Narratives that associate politicians with specific ethnic groups and label them in negative ways. Elections are viewed through the lens of a competition between different ethnic groups.

\item \textbf{Ethnicity-related (Ethnic generalisation):} Narratives that generalise entire ethnic groups, treating them as homogeneous `others'.

\item \textbf{Ethnicity-related (Ethnic offensive language):} Narratives that use offensive language to describe or address individuals based on their membership in certain ethnic groups.
\item \textbf{Ethnicity-related (Threat to population narratives):} Narratives claiming that the native population is being replaced by immigrants of different ethnicities.

\item \textbf{Migration-related (Migrants societal threat):} Narratives that portray migrants as a societal threat, emphasising issues such as violence, terrorism, inequality, and crime.

\item \textbf{Distrust in institutions (Failed state):} The country is in a state of demise/collapse. Focusing on issues such as immigration, police, etc.

\item \textbf{Distrust in institutions (Criticism of national policies):} Criticising national policy positions on contentious issues like pension, military conscription, and wealth disparity.
\item \textbf{Distrust in democratic system (Elections are rigged):} Narratives alleging election manipulation through various means, including political interference, exclusion, anti-vote measures, lack of transparency, and distrust in electoral institutions or polling agencies.

\item \textbf{Distrust in democratic system (Anti-Political system):} Narratives critical of the political system, asserting that political parties are neither beneficial nor necessary for a well-functioning country.

\item \textbf{Distrust in democratic system (Anti-Media):} Narratives critical of the media, including attacks on journalists, allegations of corruption or complicity in undermining the country, lack of balance and impartiality, and claims of favoritism towards a particular political faction.

\item \textbf{Distrust in democratic system (Immigrants right to vote):} Narratives surrounding immigrants' right to vote that claim their participation biases election outcomes or that they are deliberately excluded from the process, with potential impacts on election fairness and representation.
\item \textbf{Geopolitics (Pro-Russia):} Narratives that show Russia as strong country that is the victim of the west.

\item \textbf{Geopolitics (Foreign interference):} Narratives claiming that foreign actors are meddling in the affairs of EU countries.

\item \textbf{Geopolitics (Anti-international institutions):} Narratives that oppose or criticise international organisations such as NATO.
\item \textbf{Anti-Elites (Soros):} Narratives targeting George Soros, accusing him of manipulating political processes.
\item \textbf{Anti-Elites (World Economic Forum / Great Reset):}  Narratives claiming that the World Economic forum has/seeks control over society.

\item \textbf{Anti-Elites (Antisemitism):}  Narratives accusing Jews or Jewish organisations of having disproportionate power and influence over society and politics.

\item \textbf{Anti-Elites (Green Agenda):} Narratives that criticise the ``Green Agenda'' and green activism.

\end{enumerate}

\section{Annotation Methodology}
\label{apx:annotation-method}

\subsection{Screenshots from the Annotation Framework}

\begin{figure}[!h]

\centering
\includegraphics[width=0.48\textwidth]{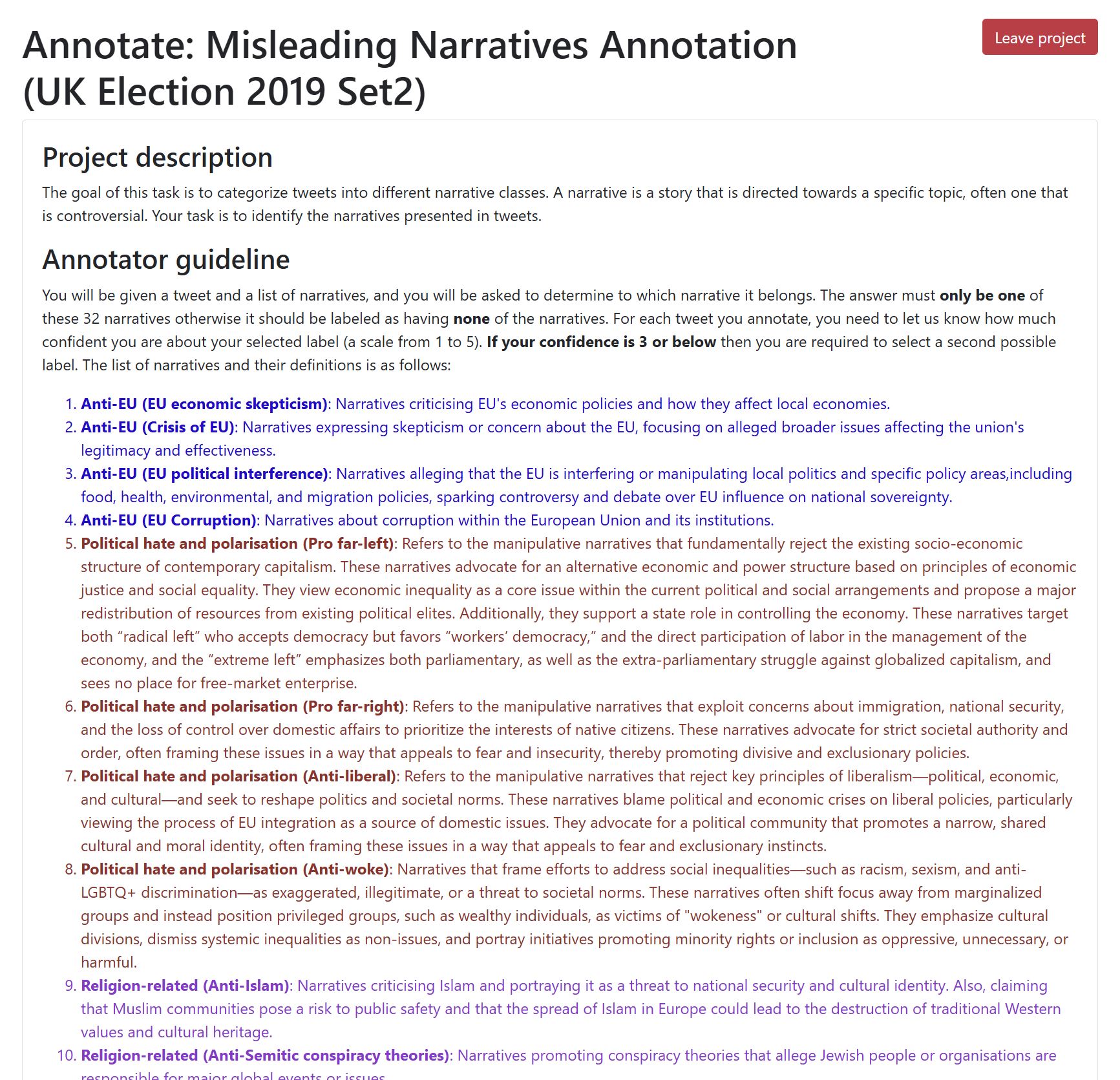} 
\caption{Annotation guidelines.}
\label{fig:guidelines}
\end{figure}

\begin{figure}[!t]
\centering
\includegraphics[width=0.48\textwidth]{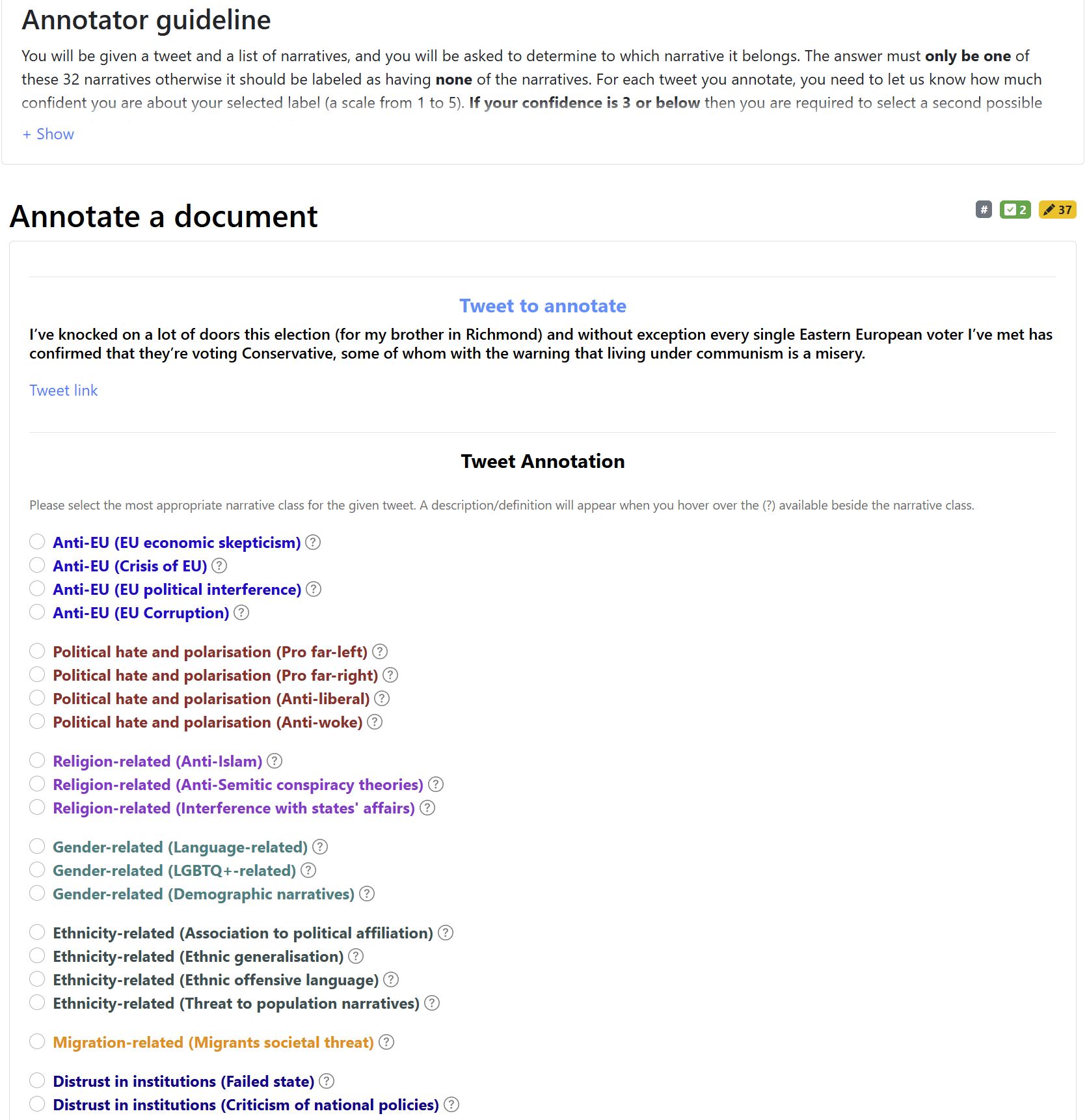} 
\caption{Selecting the first choice during annotation.}
\label{fig:annotation1}
\end{figure}

\begin{figure}[!t]
\centering
\includegraphics[width=0.48\textwidth]{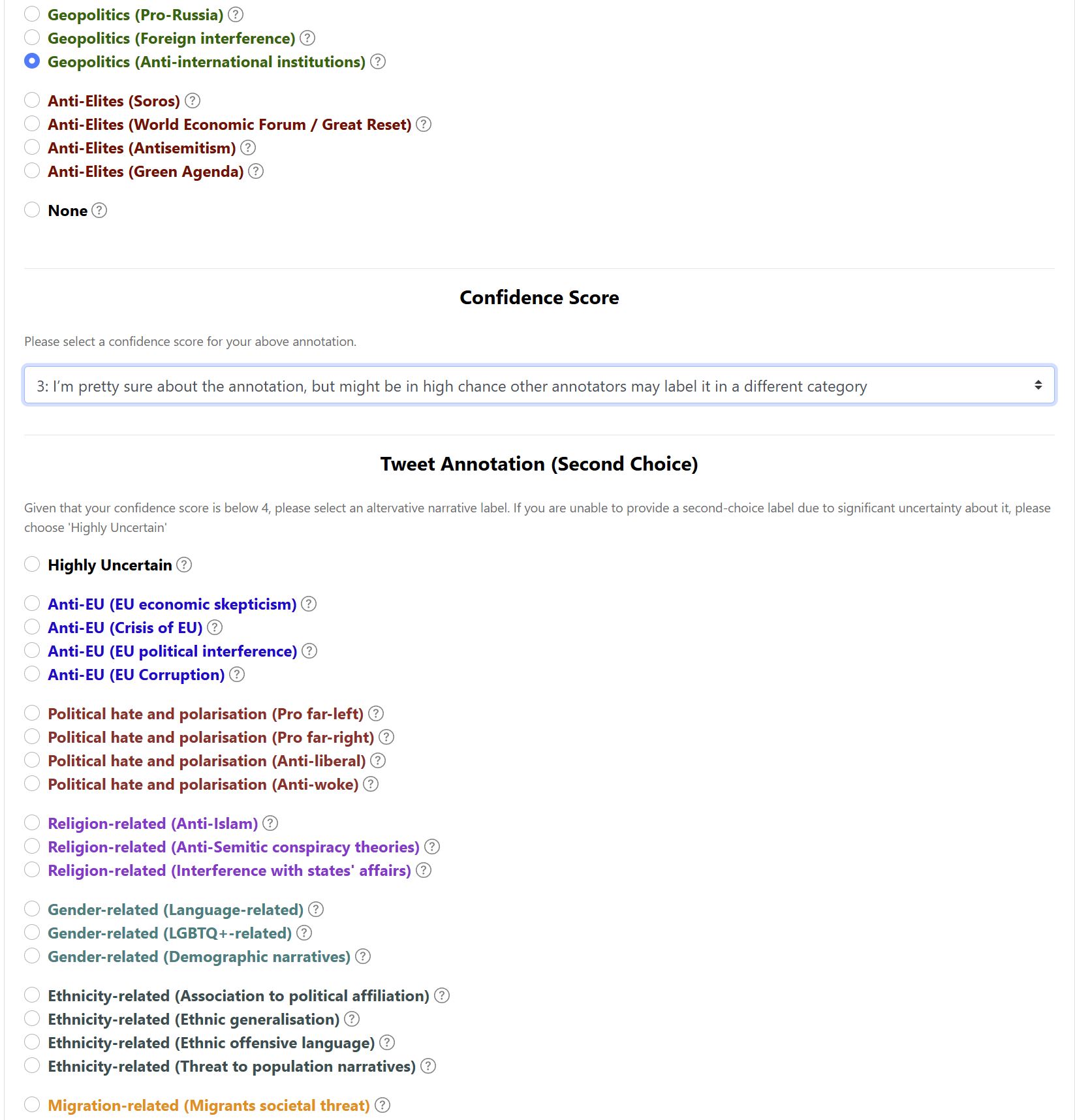} %
\caption{Selecting the second choice during annotation.}
\label{fig:annotation2}
\end{figure}

\begin{figure}[h]
\centering
\includegraphics[width=0.48\textwidth]{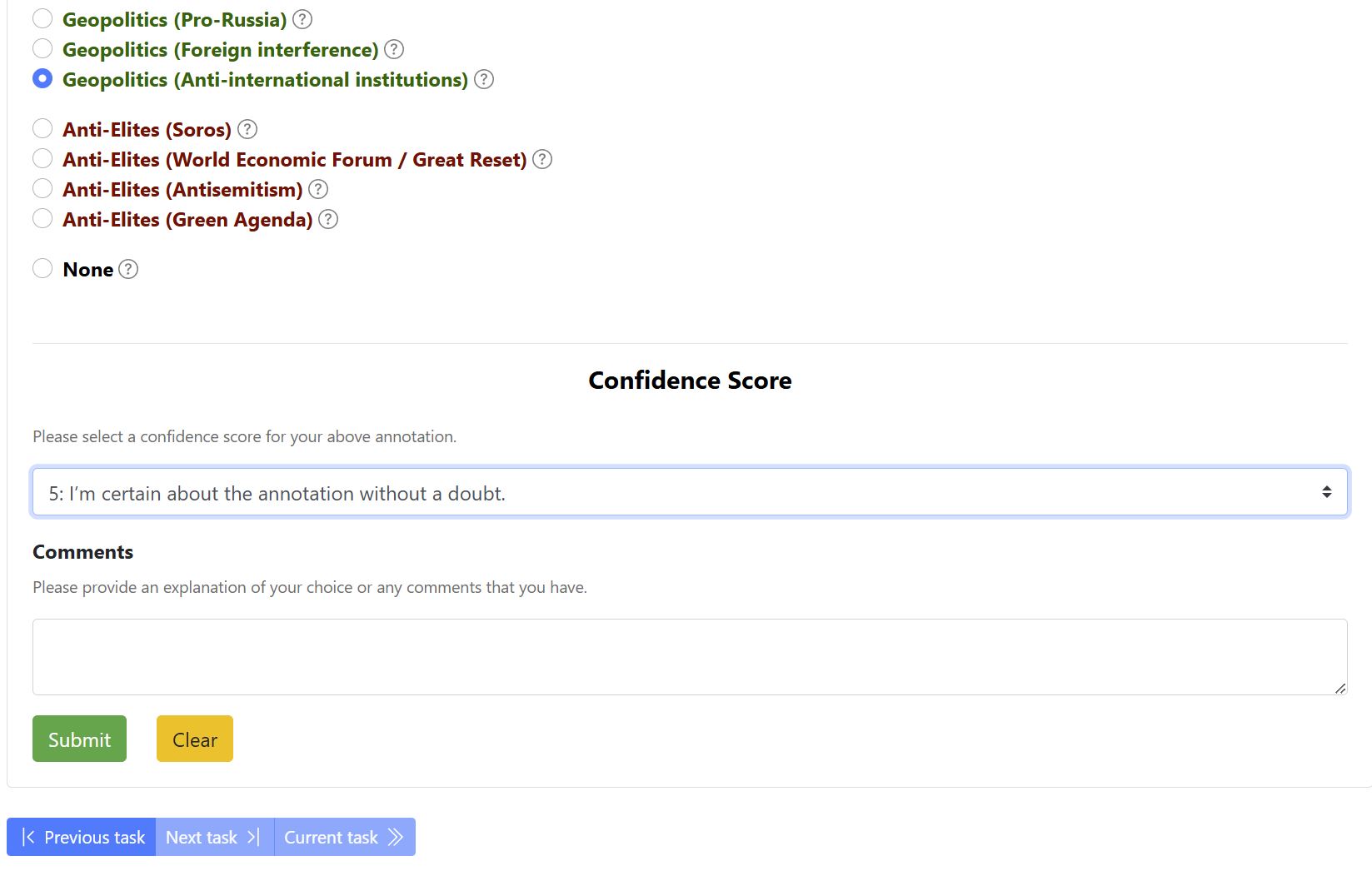} %
\caption{A second choice was not required if the confidence score is above 3.}
\label{fig:confidence}
\end{figure}

\begin{figure}[!htb]
\centering
\includegraphics[width=0.48\textwidth]{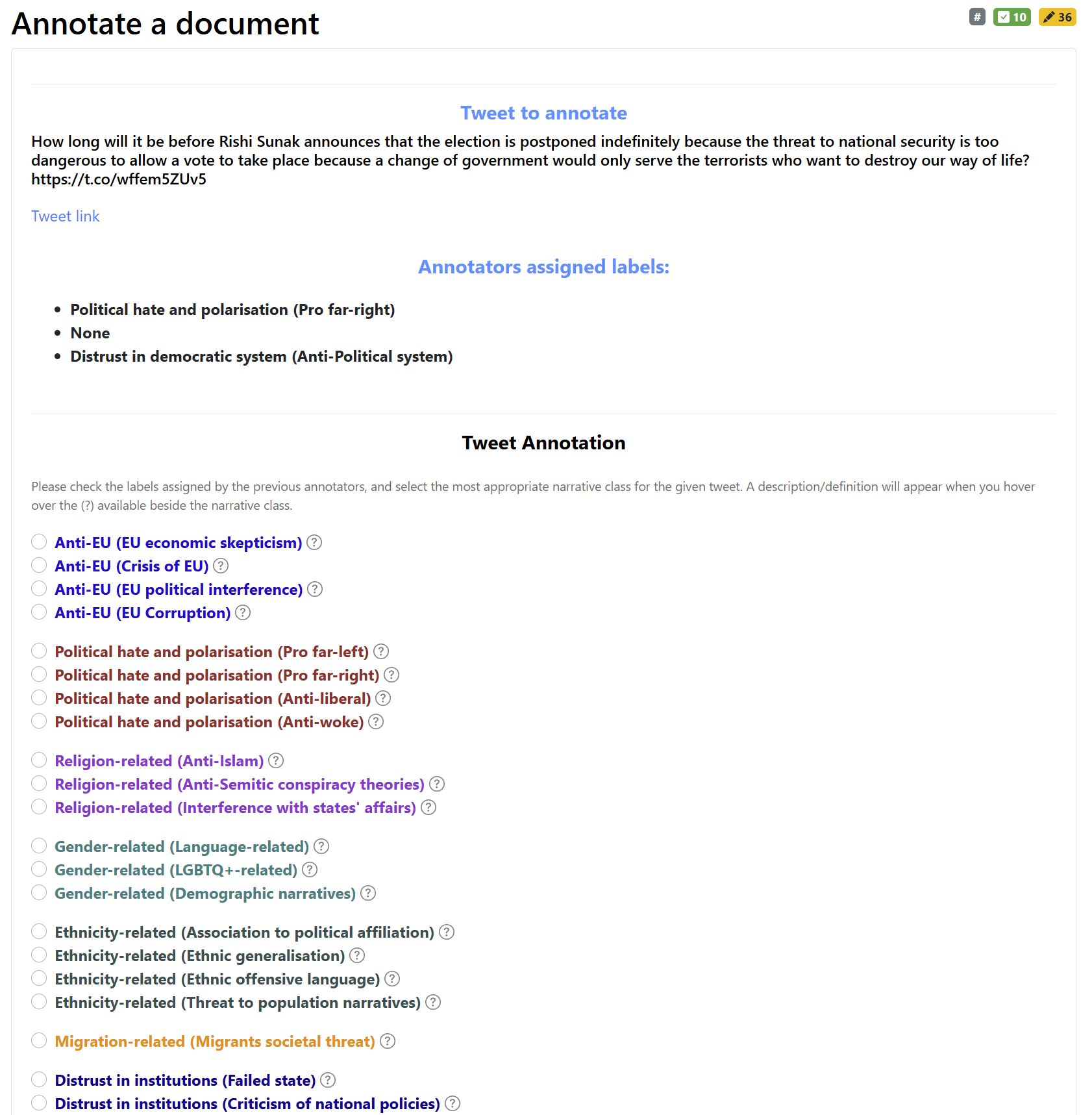} 
\caption{Stage 3 of the annotation process (consolidation).}
\label{fig:consolidation}
\end{figure}

We present in this section some screenshots from our annotation framework in Figure~\ref{fig:guidelines}, \ref{fig:annotation1}, \ref{fig:annotation2}, \ref{fig:confidence}, and Figure~\ref{fig:consolidation}. It is worth noting, that the annotators attended a training session besides being given the annotation guidelines. These session aimed to familiarise them with the task and included the presentation of example tweets for better understanding.

\subsection{Annotator Confidence Scores}

The following confidence scores descriptions were presented to the annotators:
\begin{itemize}
    \item \textbf{Score 1:} I'm really unsure about the annotation. It may belong to another category as well, you may wish to discard this instance from the training.
    \item \textbf{Score 2:} I'm not sure about the annotation, it seems it also belongs to other categories.
    \item \textbf{Score 3:} I'm pretty sure about the annotation, but might be in high chance other annotators may label it in a different category
    \item \textbf{Score 4:} I'm confident about the annotation, but might be in small chance other annotators may label it in a different category.
    \item \textbf{Score 5:} I'm certain about the annotation without a doubt.
\end{itemize}

\section{Prompt Template for the pre-annotation annotation task using LLMs}
\label{apx:prompts-llm-annot}

\noindent{\texttt{You are a social media moderator that will determine the narratives in tweets posted by users. \\
You will be given a single tweet and a list of narratives, and you will be asked to determine to which narrative it belongs.\\
The answer must only be one of these narratives [Anti-EU (EU economic scepticism), Anti-EU (Crisis of EU), Anti-EU (EU political interference), Anti-EU (EU Corruption), Political hate and polarisation (Pro far-left), Political hate and polarisation (Pro far-right), Political hate and polarisation (Anti-liberal), Political hate and polarisation (Anti-woke), Religion-related (Anti-Islam),  Religion-related (Anti-Semitic conspiracy theories),  Religion-related (Interference with states' affairs), Gender-related (Language-related), Gender-related (LGBTQ+-related), Gender-related (Demographic narratives), Ethnicity-related (Association to political affiliation), Ethnicity-related (Ethnic generalisation), Ethnicity-related (Ethnic offensive language), Ethnicity-related (Threat to population narratives), Migration-related (Migrants societal threat), Distrust in institutions (Failed state), Distrust in institutions (Criticism of national policies), Distrust in democratic system (Elections are rigged)', Distrust in democratic system (Anti-Political system), Distrust in democratic system (Anti-Media), Distrust in democratic system (Immigrants right to vote), Geopolitics (Pro-Russia), Geopolitics (Foreign interference), Geopolitics (Anti-international institutions), Anti-Elites (Soros), Anti-Elites (World Economic Forum / Great Reset), Anti-Elites (Antisemitism), Anti-Elites (Green Agenda), None]\\
 You should not provide any explanation or justification.\\
 Question: Can you detect the narrative in this tweet given the list of narratives presented to you?\\
 Tweet: ``Tweet''}}

\section{Prompt Template for Topic Labelling (Topic Modelling)}
\label{apx:prompts-topics}

\texttt{I have a topic that contains the following 
documents:
[DOCUMENTS]\\
The topic is described by the following keywords: [KEYWORDS]\\
Based on the information above, extract a short topic label in the following format:
topic: ``short topic label''}

\section{Prompt Template for Narrative Detection}

\subsection{Prompt Template for Narrative Detection (zero shot)}
\noindent{\texttt{You are a content moderator who will monitor if there are any misleading narratives in the tweets posted by users.\\
		You will be given a single tweet and a list of narratives, and your job is to read the tweet and determine if it is expressing any one of the narratives listed to you.\\
        The answer must only be one of these narratives [Anti-EU (EU economic scepticism), Anti-EU (Crisis of EU), Anti-EU (EU political interference), Anti-EU (EU Corruption), Political hate and polarisation (Pro far-left), Political hate and polarisation (Pro far-right), Political hate and polarisation (Anti-liberal), Political hate and polarisation (Anti-woke), Religion-related (Anti-Islam),  Religion-related (Anti-Semitic conspiracy theories),  Religion-related (Interference with states' affairs), Gender-related (Language-related), Gender-related (LGBTQ+-related), Gender-related (Demographic narratives), Ethnicity-related (Association to political affiliation), Ethnicity-related (Ethnic generalisation), Ethnicity-related (Ethnic offensive language), Ethnicity-related (Threat to population narratives), Migration-related (Migrants societal threat), Distrust in institutions (Failed state), Distrust in institutions (Criticism of national policies), Distrust in democratic system (Elections are rigged)', Distrust in democratic system (Anti-Political system), Distrust in democratic system (Anti-Media), Distrust in democratic system (Immigrants right to vote), Geopolitics (Pro-Russia), Geopolitics (Foreign interference), Geopolitics (Anti-international institutions), Anti-Elites (Soros), Anti-Elites (World Economic Forum / Great Reset), Anti-Elites (Antisemitism), Anti-Elites (Green Agenda), None]\\
         You should not provide any explanation or justification.\\
         Question: Can you detect any misleading narrative in this tweet given the list of narratives presented to you?\\
         Tweet:``Tweet''}}

\subsection{Prompt Template for Narrative Detection (few shot)}
\noindent{\texttt{You are a content moderator who will monitor if there are any misleading narratives in the tweets posted by users.\\
		You will be given a single tweet and a list of narratives, and your job is to read the tweet and determine if it is expressing any one of the narratives listed to you.\\
        The answer must only be one of these narratives [Anti-EU (EU economic scepticism), Anti-EU (Crisis of EU), Anti-EU (EU political interference), Anti-EU (EU Corruption), Political hate and polarisation (Pro far-left), Political hate and polarisation (Pro far-right), Political hate and polarisation (Anti-liberal), Political hate and polarisation (Anti-woke), Religion-related (Anti-Islam),  Religion-related (Anti-Semitic conspiracy theories),  Religion-related (Interference with states' affairs), Gender-related (Language-related), Gender-related (LGBTQ+-related), Gender-related (Demographic narratives), Ethnicity-related (Association to political affiliation), Ethnicity-related (Ethnic generalisation), Ethnicity-related (Ethnic offensive language), Ethnicity-related (Threat to population narratives), Migration-related (Migrants societal threat), Distrust in institutions (Failed state), Distrust in institutions (Criticism of national policies), Distrust in democratic system (Elections are rigged)', Distrust in democratic system (Anti-Political system), Distrust in democratic system (Anti-Media), Distrust in democratic system (Immigrants right to vote), Geopolitics (Pro-Russia), Geopolitics (Foreign interference), Geopolitics (Anti-international institutions), Anti-Elites (Soros), Anti-Elites (World Economic Forum / Great Reset), Anti-Elites (Antisemitism), Anti-Elites (Green Agenda), None]\\
        Here are some example tweets and their associated narrative label to help you decide:\\
        ...\\
        ...\\
        Tweet: Brussels chaos: Spain follows Poland in shock threat to quit EU 'No more humiliation!'   The EU are finished and I'm glad to say we started the destruction of it.  Europe will be a better place without it\\
        Narrative:  Anti-EU (Crisis of EU)\\
        ...\\
        Tweet: Scary scenes on London streets. Every day, UK and London in particular, look like a city from the Middle East, with angry wild Islamists bullying and threatening people!   Sick and scary!  Shame on @MayorofLondon'\\
        Narrative:  Religion-related (Anti-Islam)\\
        ...\\
        You should not provide any explanation or justification.\\
        Question: Can you detect any misleading narrative in this tweet given the list of narratives presented to you?\\
        Tweet:``Tweet''}
\subsection{Prompt Template for Narrative Detection (zero shot) with Narrative Descriptions}}
\noindent{\texttt{You are a content moderator who will monitor if there are any misleading narratives in the tweets posted by users.\\
		You will be given a single tweet and a list of narratives and their descriptions, and your job is to read the tweet and determine if it is expressing any one of the narratives listed to you.\\
        The answer must only be one of these narratives [Anti-EU (EU economic scepticism), Anti-EU (Crisis of EU), Anti-EU (EU political interference), Anti-EU (EU Corruption), Political hate and polarisation (Pro far-left), Political hate and polarisation (Pro far-right), Political hate and polarisation (Anti-liberal), Political hate and polarisation (Anti-woke), Religion-related (Anti-Islam),  Religion-related (Anti-Semitic conspiracy theories),  Religion-related (Interference with states' affairs), Gender-related (Language-related), Gender-related (LGBTQ+-related), Gender-related (Demographic narratives), Ethnicity-related (Association to political affiliation), Ethnicity-related (Ethnic generalisation), Ethnicity-related (Ethnic offensive language), Ethnicity-related (Threat to population narratives), Migration-related (Migrants societal threat), Distrust in institutions (Failed state), Distrust in institutions (Criticism of national policies), Distrust in democratic system (Elections are rigged)', Distrust in democratic system (Anti-Political system), Distrust in democratic system (Anti-Media), Distrust in democratic system (Immigrants right to vote), Geopolitics (Pro-Russia), Geopolitics (Foreign interference), Geopolitics (Anti-international institutions), Anti-Elites (Soros), Anti-Elites (World Economic Forum / Great Reset), Anti-Elites (Antisemitism), Anti-Elites (Green Agenda), None]\\
        Please check the following narrative descriptions before making a decision:\\
        Anti-EU (EU economic scepticism): Narratives criticising EU's economic policies and how they affect local economies.\\
        Anti-EU (Crisis of EU): Narratives expressing scepticism or concern about the EU, focusing on alleged broader issues affecting the union's legitimacy and effectiveness.\\
         Anti-EU (EU political interference): Narratives alleging that the EU is interfering or manipulating local politics and specific policy areas,including food, health, environmental, and migration policies, sparking controversy and debate over EU influence on national sovereignty.\\
         ...\\
         None: None of the above narratives.\\
         You should not provide any explanation or justification.\\
         Question: Can you detect any misleading narrative in this tweet given the list of narratives presented to you?\\
         Tweet:``Tweet''}}
\subsection{Prompt Template for Narrative Detection (few shot) with Narrative Descriptions}
\texttt{You are a content moderator who will monitor if there are any misleading narratives in the tweets posted by users.\\
		You will be given a single tweet and a list of narratives and their descriptions, and your job is to read the tweet and determine if it is expressing any one of the narratives listed to you.\\
        The answer must only be one of these narratives [Anti-EU (EU economic scepticism), Anti-EU (Crisis of EU), Anti-EU (EU political interference), Anti-EU (EU Corruption), Political hate and polarisation (Pro far-left), Political hate and polarisation (Pro far-right), Political hate and polarisation (Anti-liberal), Political hate and polarisation (Anti-woke), Religion-related (Anti-Islam),  Religion-related (Anti-Semitic conspiracy theories),  Religion-related (Interference with states' affairs), Gender-related (Language-related), Gender-related (LGBTQ+-related), Gender-related (Demographic narratives), Ethnicity-related (Association to political affiliation), Ethnicity-related (Ethnic generalisation), Ethnicity-related (Ethnic offensive language), Ethnicity-related (Threat to population narratives), Migration-related (Migrants societal threat), Distrust in institutions (Failed state), Distrust in institutions (Criticism of national policies), Distrust in democratic system (Elections are rigged)', Distrust in democratic system (Anti-Political system), Distrust in democratic system (Anti-Media), Distrust in democratic system (Immigrants right to vote), Geopolitics (Pro-Russia), Geopolitics (Foreign interference), Geopolitics (Anti-international institutions), Anti-Elites (Soros), Anti-Elites (World Economic Forum / Great Reset), Anti-Elites (Antisemitism), Anti-Elites (Green Agenda), None]\\
        Please check the following narrative descriptions before making a decision:\\
        Anti-EU (EU economic scepticism): Narratives criticising EU's economic policies and how they affect local economies.\\
         ...\\
        None: None of the above narratives.\\
        Here are some example tweets and their associated narrative label to help you decide:\\
        ...\\
        Tweet: Scary scenes on London streets. Every day, UK and London in particular, look like a city from the Middle East, with angry wild Islamists bullying and threatening people!   Sick and scary!  Shame on @MayorofLondon'\\
        Narrative:  Religion-related (Anti-Islam)\\
        ...\\
        You should not provide any explanation or justification.\\
        Question: Can you detect any misleading narrative in this tweet given the list of narratives presented to you?\\
        Tweet:``Tweet''}

\section{Topic Modeling}
\label{apx:topic-models}

\begin{figure*}[h]
\centering
\includegraphics[width=0.7\textwidth]{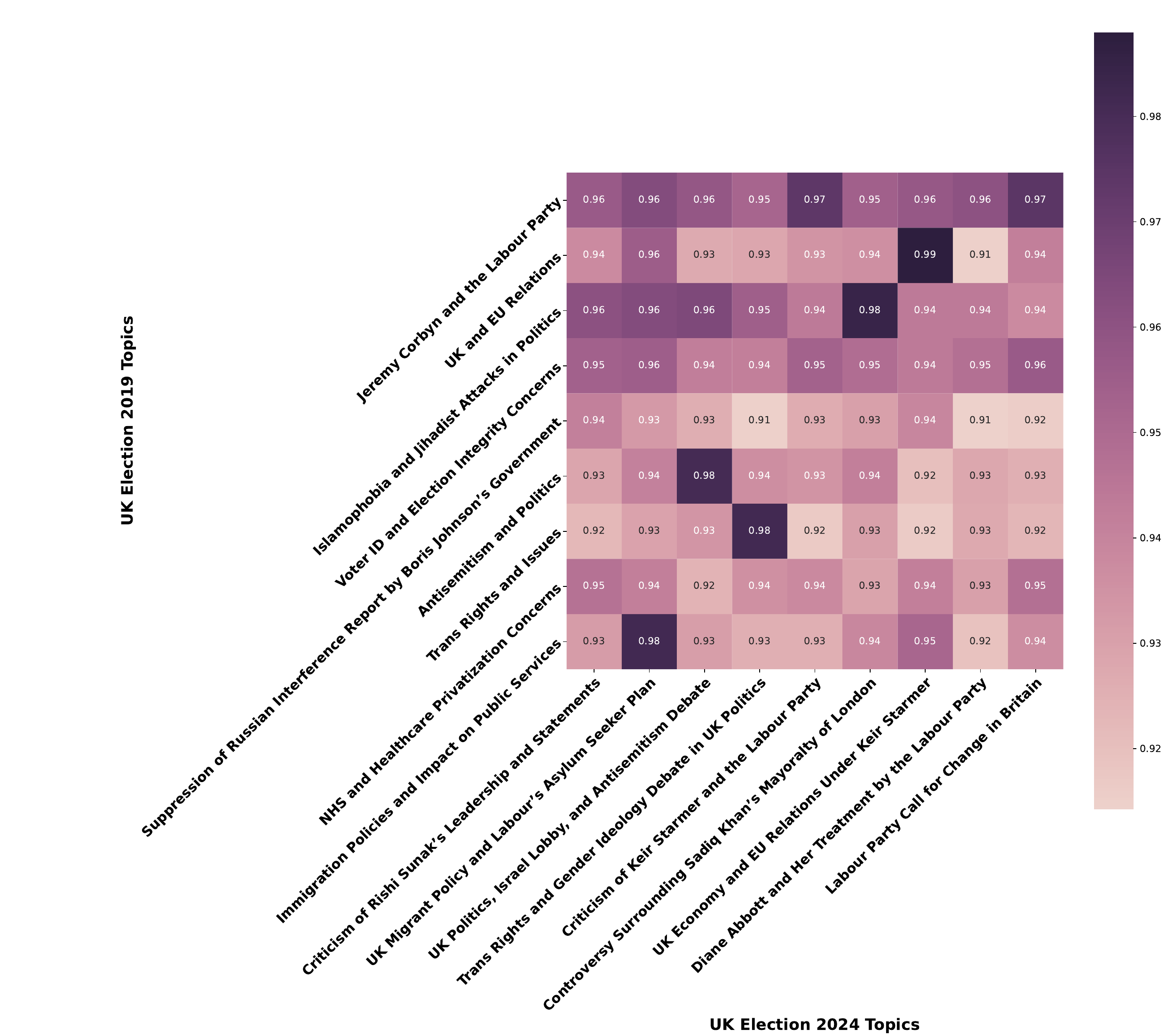} 
\caption{Cosine Similarity Matrix for Election 2019 and 2024 Topics Similarity.}
\label{fig:topics_heatmap}
\end{figure*}

In Figure~\ref{fig:topics_heatmap}, we present the topic similarity across both 2019 and 2024 elections.

\section{Data Statistics and Examples: Further Details}
\label{apx:dataset-further-stats}

This section presents further details about the UKElectionNarratives dataset. Table~\ref{tab:labels_distribution} shows the label distribution in the dataset's train, development, and test subsets 
 which were used in our benchmark experiments. Table~\ref{fig:example} presents example tweets from the dataset.

\section{Computational Resources}
\label{apx:computation-resources}

To perform our experiments, we used our internal servers using two different GPUs specifically  \textit{NVIDIA A100-PCIE-40GB} and \textit{NVIDIA RTX A4500}.

\begin{table}[!h]
  \tiny
  \centering
  \begin{tabular}{l|ccc}
    \toprule
    &\bf Train &\bf Dev &\bf Test\\\midrule
    \bf None&548&76& 157\\
    \bf Distrust in institutions (Criticism of national policies) &171&23& 47\\

    \bf Gender-related (LGBTQ+-related) &119&17&33\\
    \bf Religion-related (Anti-Islam) &72&10&19\\
    \bf Distrust in democratic system (Anti-Media) &72&10&19\\
    \bf Migration-related (Migrants societal threat) &65&9&17\\
     \bf Geopolitics (Foreign interference)  &57&8&16\\
    \bf Distrust in democratic system (Elections are rigged) &50&7&14\\ 
        \bf Distrust in institutions (Failed state) &45&6& 13\\
    \bf     Political hate and polarisation (Pro far-left)  &39&5&12\\ 
    \bf Political hate and polarisation (Pro far-right) &38&5&11\\ 
   \bf  Anti-EU (EU political interference)   &23&3&7\\
     \bf  Anti-EU (Crisis of EU)     &17&2&6\\
    \bf  Anti-EU (EU economic scepticism)  &13&2&4\\
     \bf  Distrust in democratic system (Anti-Political system)  &13&1&4\\ 
    \bf  Political hate and polarisation (Anti-liberal) &12&1&4\\
    \bf  Political hate and polarisation (Anti-woke)  &6&1&1\\ 
    \bf  Ethnicity-related (Association to political affiliation)  &5&1&1\\
    \bf Gender-related (Language-related)   &5&1&1\\ 
    \bf  Ethnicity-related (Ethnic offensive language)  &4&1&1\\
    \bf Anti-Elites (Soros)   &4&1&1\\
    \bf Ethnicity-related (Threat to population narratives)   &4&1&1\\
      \bf Anti-EU (EU Corruption)  &3&1&1\\
     \bf Geopolitics (Pro-Russia)   &3&1&1\\
\bf Religion-related (Anti-Semitic conspiracy theories)   &3&1&1\\
  \bf Ethnicity-related (Ethnic generalisation)  &2&1&1\\
      \bf Anti-Elites (World Economic Forum / Great Reset)    &2&1&1\\
         \bf Anti-Elites (Green Agenda)    &2&1&1\\
         \bf Geopolitics (Anti-international institutions)    &2&1&1\\
         \bf Religion-related (Interference with states' affairs)     &1&1&1\\
         \bf Anti-Elites (Antisemitism)&0&1&1\\
         \bf Gender-related (Demographic narratives)&0&0&1\\
          \bf Distrust in democratic system (Immigrants right to vote)&0&0&1\\
    \midrule
    \bf Total &1400&200&400\\
    \midrule
    \bf Election 2019 tweets &700&100&200\\
    \bf Election 2024 tweets&700&100&200\\
 \bottomrule   
\end{tabular}
 \caption{Data Splits and Label Distributions. }

  \label{tab:labels_distribution}
\end{table}

\begin{table*}[!h]
\centering

\begin{tabular}{p{12cm}}\toprule

\textbf{Tweet:} The process for casting a vote in UK elections is nothing short of archaic.
No ID.
No CCTV of who is coming \& going.
Who watches the boxes when they are driven to the central place for counting?
Pencils!?

The whole procedure is ripe for fraud \& must change.

\#GE2109\\ 

\textbf{Super-narrative:} Distrust in democratic system\\
 \textbf{Narrative:} Elections are rigged \\

\midrule

\textbf{Tweet:} As Rishi Sunak brags about his plans for prisons \#PMQs  Ministers today announced emergency measures to keep defendants in police cells as PRISONS HAVE RUN OUT OF SPACE   What living political TORY hell is this?\\ 

\textbf{Super-narrative:} Distrust in institutions \\
 \textbf{Narrative:} Failed state \\

\midrule

\textbf{Tweet:} To all young people grappling with rising house costs...   Mass immigration has caused this.   Demand has outstripped supply.  We will NEVER be able to build enough houses to support an extra 400,000 people EACH YEAR.  In reality, freedom of movement has been a disaster for you.\\ 

\textbf{Super-narrative:} Migration-related\\
 \textbf{Narrative:} Migrants societal threat \\
\midrule
\textbf{\textbf{Tweet:}} The Greens should be reminded the UK accounts for 1\% of global CO2 emissions, whilst their policies would inflict major economic damage.\\ 

\textbf{Super-narrative:} Anti-elites\\
 \textbf{Narrative:} Green Agenda \\

\bottomrule
\\
\end{tabular}
\caption{Sample tweets and corresponding super-narrative and narrative from \textbf{UKElectionNarratives}. }
\label{fig:example}
\end{table*}

\end{document}